%% file: main.tex
\documentclass[10pt,twocolumn,letterpaper]{article}

\usepackage[pagenumbers]{cvpr}      

\input{preamble}

\usepackage{multicol}
\usepackage{afterpage}
\usepackage{float}
\usepackage{multirow}
\usepackage{bbding} 
\usepackage{colortbl} 
\usepackage{xcolor} 
\usepackage{array} 
\usepackage[bottom]{footmisc} 
\definecolor{cvprblue}{rgb}{0.21,0.49,0.74}
\usepackage[pagebackref,breaklinks,colorlinks,citecolor=cvprblue]{hyperref}
\definecolor{lightgray}{gray}{0.9}
\def\paperID{4656} 
\def\confName{CVPR}
\def\confYear{2025}

\title{MaSS13K: A Matting-level Semantic Segmentation Benchmark}

\author{Chenxi Xie\textsuperscript{1,2,\dag} ~~~~~~~~Minghan Li\textsuperscript{1,2,\dag}  ~~~~~~~~Hui Zeng\textsuperscript{2} ~~~~~~~~Jun Luo\textsuperscript{2} ~~~~~~~~Lei Zhang\textsuperscript{1,2*}
\\
\textsuperscript{1}The Hong Kong Polytechnic University ~~~~\textsuperscript{2}OPPO Research Institute
}

\let\oldtwocolumn\twocolumn
\renewcommand\twocolumn[1][]{%
    \oldtwocolumn[{#1}{
    \begin{center}
           \includegraphics[width=0.9\textwidth]{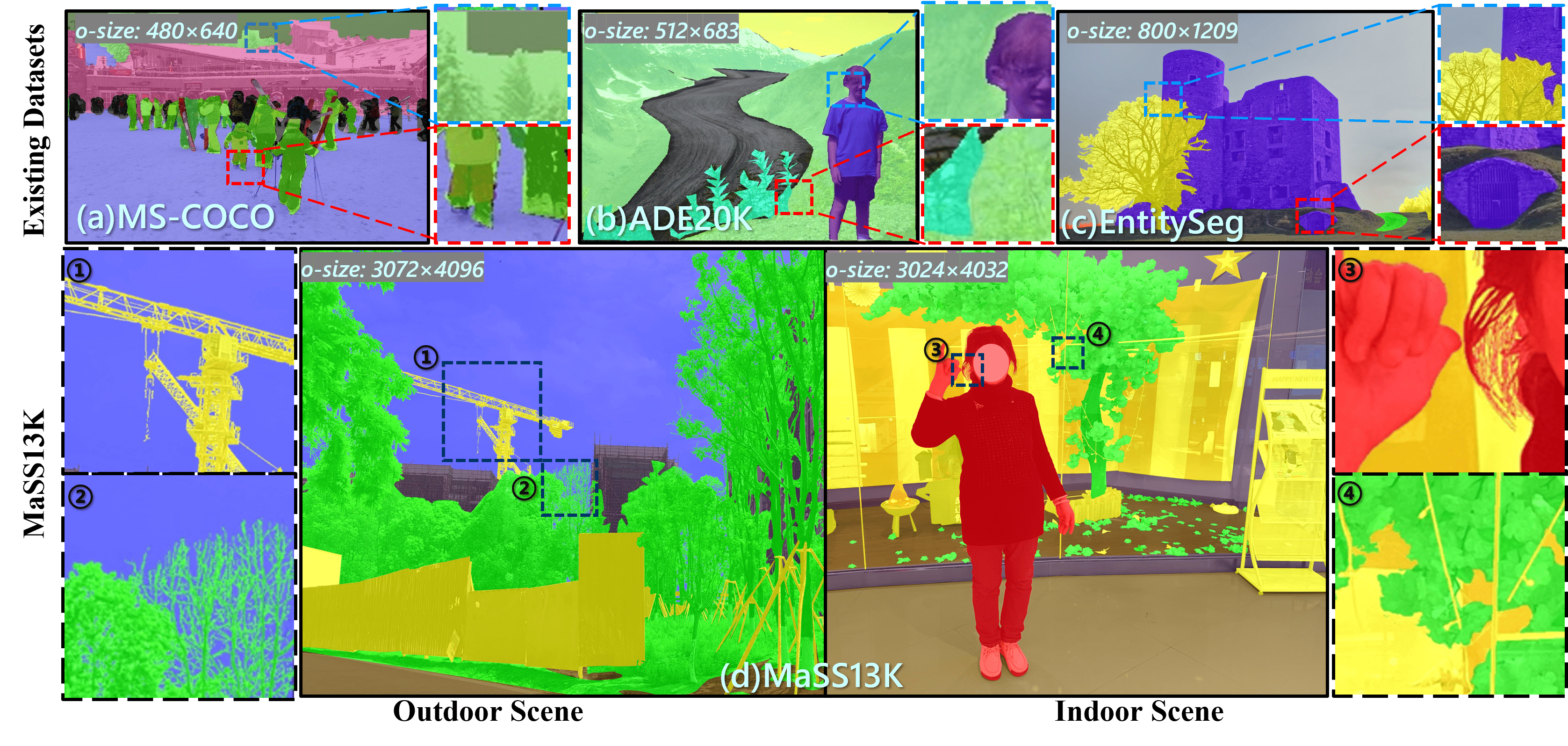}
           \vspace{-4mm}
           \captionof{figure}{Comparisons on the resolution and annotation quality between existing semantic segmentation datasets (a) COCO, (b) ADE20K, (c) EntitySeg and (d) our established dataset MaSS13K. Note that the `others' category, which is highlighted in yellow, in our MaSS13K actually contains a group of well-segmented objects without specific class names. Please zoom-in for a better view.}
           \label{fig:show_result}
        \end{center}
    }]
}

\begin{document}

\maketitle

\input{sec/0_abstract}    
\input{sec/1_intro}

\input{sec/2_related}

\input{sec/3_dataset}

\input{sec/4_method}
\input{sec/5_experiments}

\input{sec/6_conclusion}


\onecolumn
\setcounter{section}{0}
\input{sec/X_suppl}
\oldtwocolumn
{
    \small
    \bibliographystyle{ieeenat_fullname}
    \bibliography{main}
}

\end{document}

%% file: preamble.tex
\usepackage[dvipsnames]{xcolor}


%% file: sec/0_abstract.tex
\renewcommand{\thefootnote}{}
\footnotetext{\textsuperscript{\dag} Equal contributions. \textsuperscript{*}Corresponding author. This work is supported by the PolyU-OPPO Joint Innovative Research Center.}
\begin{abstract}
High-resolution semantic segmentation is essential for applications such as image editing, bokeh imaging, AR/VR, etc. Unfortunately, existing datasets often have limited resolution and lack precise mask details and boundaries.
In this work, we build a large-scale, matting-level semantic segmentation dataset, named MaSS13K, which consists of 13,348 real-world images, all at 4K resolution. MaSS13K provides high-quality mask annotations of a number of objects, which are categorized into seven categories: human, vegetation, ground, sky, water, building, and others. MaSS13K features precise masks, with an average mask complexity 20-50 times higher than existing semantic segmentation datasets. We consequently present a method specifically designed for high-resolution semantic segmentation, namely MaSSFormer, which employs an efficient pixel decoder that aggregates high-level semantic features and low-level texture features across three stages, aiming to produce high-resolution masks with minimal computational cost.
Finally, we propose a new learning paradigm, which integrates the high-quality masks of the seven given categories with pseudo labels from new classes, enabling MaSSFormer to transfer its accurate segmentation capability to other classes of objects.
Our proposed MaSSFormer is comprehensively evaluated on the MaSS13K benchmark together with 14 representative segmentation models.  
We expect that our meticulously annotated MaSS13K dataset and the MaSSFormer model can facilitate the research of high-resolution and high-quality semantic segmentation. Datasets and codes can be found at \url{https://github.com/xiechenxi99/MaSS13K}. 

\end{abstract}

%% file: sec/1_intro.tex
\vspace{-0.2cm}
\section{Introduction}

Semantic segmentation, which assigns a category label to each pixel in an image, has made significant strides over the past decades, driven by the advancements of deep learning and the availability of large-scale benchmark datasets such as COCO-Stuff \cite{coco}, and ADE20K \cite{ade20k}, among others.
Building upon these datasets, researchers have developed a variety of networks, including FCN-based models \cite{fcn} such as the  DeepLab series \cite{deeplabv3,deeplabv3+}. More recently, with the introduction of transformer architectures \cite{attention,swin}, transformer-based models \cite{mask2former,maskformer} have shifted the paradigm of semantic segmentation from pixel-level classification to mask-level classification, achieving new state-of-the-art performance.

The image resolution in COCO-Stuff, ADE20K, and Pascal VOC datasets, however, is commonly below $1000 \times 1000$. As the resolution of images encountered in our daily life has significantly risen, there is a growing demand for high-resolution semantic segmentation in applications such as image editing \cite{editing}, bokeh imaging \cite{bokeh}, image retouching \cite{retouching}, AR/VR \cite{ar}, \etc., where fine-scale mask details significantly impact user experience. Recognizing the limitations of low-resolution data, researchers have started to construct some higher-resolution datasets such as Mapillary Vistas \cite{neuhold2017mapillary} and EntitySeg \cite{entityseg}. However, these still fall short with resolutions below 2K and struggle to annotate objects with complex structures and details accurately, as illustrated in \cref{fig:show_result}. Some datasets, like the DIS \cite{dis} and matting datasets \cite{aim500,matting1,matting2}, are meticulously annotated but they are designed for class-agnostic binary segmentation tasks, which can only be used to differentiate the foreground from the background, rather than parsing the entire scene into semantic categories. Therefore, there remains a demand for highly accurate semantic segmentation datasets to advance the development of high-resolution segmentation models.

To this end, we introduce the \textbf{Ma}tting-level \textbf{S}emantic \textbf{S}egmentation dataset, namely \textbf{MaSS13K}, which comprise 13,348 4K-resolution images captured from the real-world scenes. MaSS13K provides exceptionally high-quality annotations (please refer to \cref{fig:show_result}) on a variety of objects, which are categorized into seven common semantic categories, including `\texttt{human}', `\texttt{vegetation}', `\texttt{ground}', `\texttt{sky}', `\texttt{water}', `\texttt{buildings}' and `\texttt{others}'. It should be noted that \textit{`others' in MaSS13K is not the `background' class in other datasets}. It actually refers to a group of well-segmented objects such as the `crane' in \cref{fig:show_result}.
Using the mIPQ \cite{ipq} score to assess mask complexity, MaSS13K's mIPQ is 20 to 50 times higher than that of existing semantic segmentation datasets and three times higher than the finely annotated DIS dataset \cite{dis}. 
With MaSS13K, we conduct a comprehensive analysis on 14 representative semantic segmentation methods \cite{stdc,bisenet,pid,cgrseg,feed,segnext,sea,deeplabv3+,ocrnet,upernet,maskformer,mask2former,pem,mpformer}. 
We show that while these methods achieve satisfactory overall accuracy, they struggle with capturing fine boundary details and impose significant computational and memory demands for high-resolution inputs. 
There is a high demand for developing new semantic segmentation methods, which can balance the computational cost and segmentation performance in high-resolution contexts.

To tackle the above-mentioned challenges, we propose a Transformer-based model, namely MaSSFormer, specifically designed for high-resolution semantic segmentation.  
We analyze the traditional FPN-based  \cite{fpn} pixel-decoder in the context of high-resolution input, and devise a lightweight pixel-decoder to balance the computational consumption while generating high-quality masks with fine boundaries.
In specific, the decoding process is divided into two branches. In the high-level global semantic branch, we efficiently aggregate high-level semantic features across layers, providing  robust global context. Moreover, we balance detail and semantic information by expanding the receptive field to capture multi-scale structures. 
In the low-level local structure branch, we extract edge-aware features to enhance detail segmentation accuracy. 
With the guidance of edge detection, we fuse the features in the two branches and generate the final feature with accurate details.
Extensive experiments on MaSS13K dataset demonstrate that our proposed MaSSFormer outperforms existing semantic segmentation methods, particularly in boundary quality.

While the MaSS13K dataset contains seven classes, its highly detailed semantic annotations enable networks to learn the general ability to accurately segment regions based on class-agnostic boundaries. To validate this point, we develop a simple yet effective training pipeline that uses powerful pre-trained models to generate low-quality pseudo-labels for new classes, which are then combined with our finely annotated labels to train MaSSFormer. Our experiments show that MaSSFormer can produce higher-quality masks for these newly introduced classes, effectively transferring its fine segmentation abilities to novel categories. This highlights the significant potential of the meticulously annotated MaSS13K dataset for advancing future research in high-resolution and high-quality semantic segmentation.

Our contributions are summarized as follows:
\begin{itemize}
    \item We introduce MaSS13K, a large-scale semantic segmentation dataset, containing 13K 4K-resolution images with ultra-high quality semantic annotations.
    \item We propose a simple yet effective baseline MaSSFormer for high-resolution semantic segmentation, which surpasses existing methods, particularly in boundary quality.
    \item We devise a pipeline to empower the model with the capacity to accurately segment novel classes of objects without extra human efforts, revealing the potential of the high-quality semantic annotations in MaSS13K for future high-resolution segmentation research.
     \item We conduct a comprehensive benchmarking on the MaSS13k dataset by evaluating MaSSFormer with 14 cutting-edge semantic segmentation methods.
\end{itemize}

%% file: sec/2_related.tex

\begin{figure}[t]
    \centering
    \includegraphics[width=0.95\linewidth]{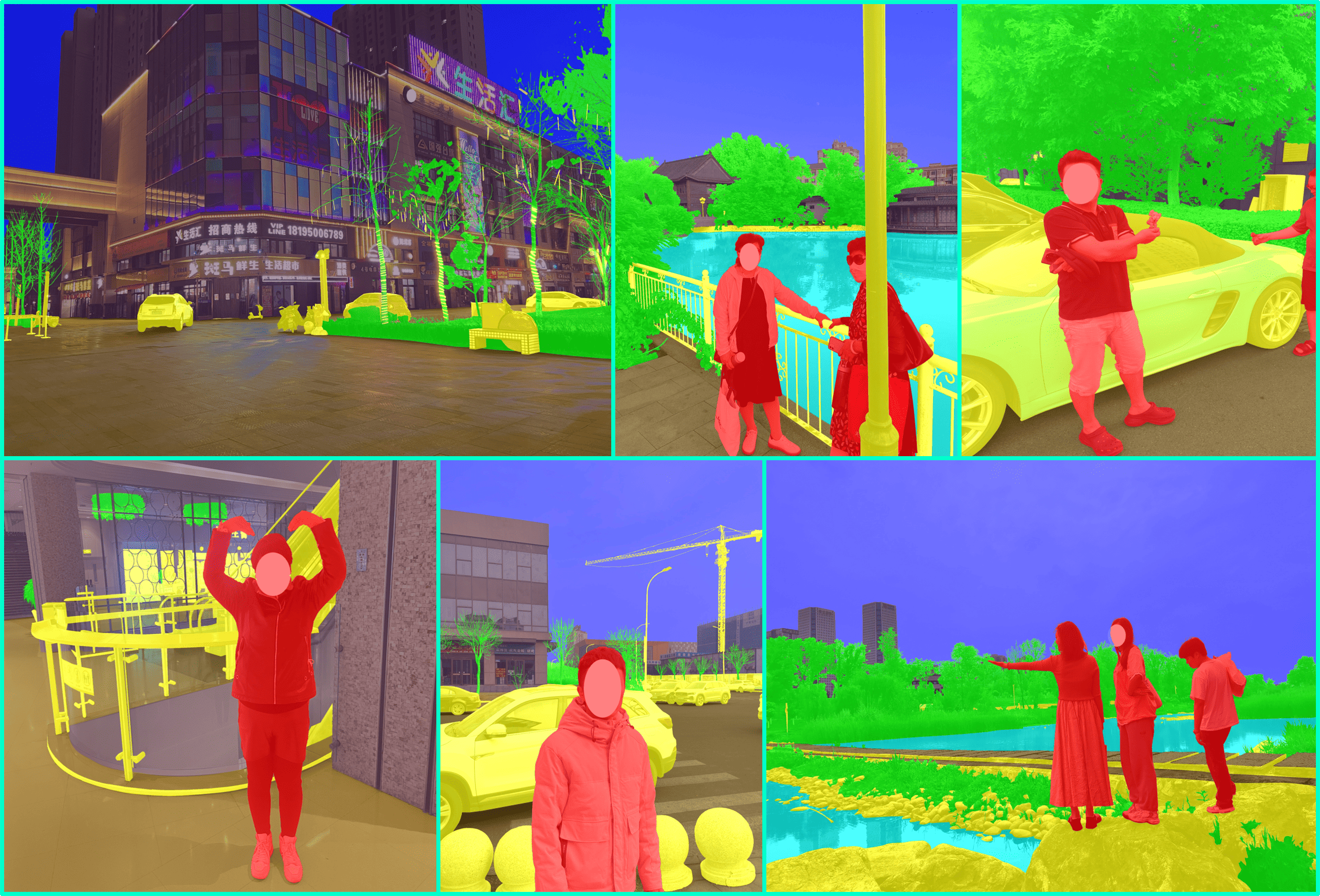}
    \caption{Some typical scenes in our MaSS13K dataset.}
    \label{fig:data_stat}
    \vspace{-3mm}
\end{figure}
\section{Related Work}
\label{sec:related_work}

\label{sec:dataset}
\textbf{Image Segmentation Datasets}.
Numerous segmentation datasets have been developed recently, generally falling into two categories: multi-class and binary segmentation datasets. Multi-class datasets are primarily designed for tasks such as semantic segmentation \cite{semantic}, instance segmentation \cite{instance}, and panoptic segmentation \cite{panoptic}. They focus on high-level semantics within scenes and are typically large in scale, encompassing a variety of semantic categories, such as COCO \cite{coco} and ADE20K \cite{ade20k}, EntitySeg \cite{entityseg}, Cityscapes \cite{cityscapes} and others \cite{pascal,neuhold2017mapillary}. However, these datasets often suffer from lower resolution and poor annotation quality.
On the other hand, binary segmentation datasets focus on tasks like salient object detection (SOD) \cite{uhrsd,hrsod}, camouflaged object detection (COD) \cite{fri,fan2020camouflaged}, and image matting \cite{aim500}. They emphasize low-level structures of target objects with fine annotations but are class-agnostic and object-centric, making them unsuitable for full scene parsing.
Our MaSS13K addresses these limitations and integrates the strengths of both types of datasets. It includes a  range of semantic categories, providing comprehensive coverage of entire scenes. Additionally, it offers high-resolution and matting-level annotations, surpassing the accuracy of existing datasets. 

\noindent\textbf{Semantic Segmentation Methods}.
With the development of deep learning, semantic segmentation has made significant progress. Initially, FCN-based \cite{fcn} methods dominate the field, with notable methods including DeepLab series \cite{deeplabv1,deeplabv2,deeplabv3}, PSPNet \cite{pspnet}, OCRNet \cite{ocrnet} and some other efficient methods \cite{pid,bisenet}. These methods introduce various modules to aggregate multi-scale semantic information, classifying each pixel directly. With the rise of Transformers and their  application in object detection \cite{detr}, unified mask classification frameworks such as MaskFormer \cite{maskformer,mask2former,pem} have been proposed for multiple segmentation tasks. This approach decouples mask prediction from classification, achieving impressive performance.
However, due to the limitations of existing semantic segmentation datasets, these methods are mostly designed for low-resolution data, resulting in challenges when applied to high-resolution datasets, such as the difficulties in global semantic awareness and poor boundary details. 
While some methods have considered edges \cite{boundary1,boundary2,pid}, they focus on how to use edges to differentiate semantic regions rather than improving boundary and detail segmentation quality.
Unlike previous methods, our MaSSFormer is designed to tackle the challenges of high-resolution semantic segmentation, ensuring semantic accuracy while producing segmentation maps with precise detail and sharp boundaries.

%% file: sec/3_dataset.tex
\section{MaSS13K Dataset}
\begin{table}[!t]
 \renewcommand{\arraystretch}{0.9}
\centering
\caption{Statistical comparison between MaSS13K and other segmentation datasets. `SS' denotes semantic segmentation and `BS' denotes binary segmentation. `\#Size' indicates the diagonal length. The values are presented as mean $\pm$ standard deviation. }
\label{tab:1}
\resizebox{\linewidth}{!}{%
\begin{tabular}{l|l|lllll}
\toprule
 Type & Dataset & \#Image &\#Size &\#Classes & \#IPQ   \\ \hline
 \multirow{4}{*}{SS}&COCO-Stuff \cite{coco}&118K &$762 \pm 70 $ &171 &$7 \pm 4 $  \\
                &ADE20K \cite{ade20k} &20K &$667 \pm 227 $ &150 &$17 \pm 10 $  \\
                &Cityscapes \cite{cityscapes}&3K &$2290 \pm 0$ &19 &$18 \pm 7$ \\
                &EntitySeg  \cite{entityseg}&11K &$2684\pm2240 $ &151 &$20 \pm 29$ \\

 \midrule
 \multirow{4}{*}{BS} &BIG  \cite{cascadepsp}&0.3K &$4655 \pm 1312$ &1 &$21 \pm 63 $ \\
 &HRSOD \cite{hrsod}&2K &$4405 \pm 1631$ &1 &$6 \pm 13$ \\
 &UHRSD \cite{uhrsd}&6K &$6185 \pm 1332$ &1 &$7 \pm 9$ \\
 &DIS5K \cite{dis}&5K &$4041 \pm 1618$  &1 &$116 \pm 452$\\
 \midrule
 \rowcolor{lightgray}SS& MaSS13K  &13K &$ 5144\pm 492 $ &7 &$383 \pm 818$ \\
 
 \bottomrule
\end{tabular}%
}
\vspace{-3mm}
\end{table}

\textbf{Data Collection.} 
To address the shortage of high-resolution semantic segmentation data, we construct a large-scale dataset, which consists of 13,348 images of 4K resolution of real-world scenes. These images were captured by various smartphones, including iPhone 14 Pro, Huawei P50, Huawei Mate50, OPPO Find X5, OPPO Reno 8, Vivo X80, and Xiaomi 12, and under different lighting conditions, weather states, and times of day.
The images encompass a diverse range of indoor and outdoor scenes, including urban areas, natural landscapes, street views, wilderness, parks, mall interiors, and other public spaces. The diverse sources of data collection not only enhance the dataset’s scene diversity but also improve the segmentation model's adaptability to different environmental conditions.
The snapshots of some typical image scenes are shown in \cref{fig:data_stat}.
Note that all images have undergone rigorous ethical and privacy reviews to ensure privacy regulations, and a screening process to remove lower quality pictures.


\noindent\textbf{Data Annotation.} 
Each image is annotated with pixel-level precision to create accurate semantic segmentation masks for seven categories: `\texttt{human}', `\texttt{water}', `\texttt{vegetation}', `\texttt{building}', `\texttt{ground}', `\texttt{sky}' and `\texttt{others}'. 
An experienced team was invited to conduct this annotation using the editing tools in PhotoShop. For each of the seven categories, annotators were asked to perform matting-level labeling, focusing on accurate boundary delineation of fine features like hair, leaf edges, and intricate object contours. Afterward, each annotated image went through a multi-stage quality control process, including peer review and final check by senior annotators, to ensure the labeling accuracy and consistency.

It should be noted that the `\texttt{others}' category in our MaSS13K dataset is different from the `background' class in other semantic segmentation datasets \cite{coco, ade20k, pascal}. In our dataset, the category `\texttt{others}' actually contains a group of accurately segmented objects but without specified class names. As shown by the yellow color in \cref{fig:show_result}, the masks of the `\texttt{others}' regions are in rich details. Therefore, they are treated as a unique class in our training and evaluation.

\noindent\textbf{Statistics Analysis.}
To better understand the statistics and advantages of our proposed MaSS13K dataset, we compare it against four popular semantic segmentation datasets, including COCO-Stuff \cite{coco}, ADE20K \cite{ade20k}, Cityscapes \cite{cityscapes}) and EntitySeg \cite{entityseg}, and four datasets used for high-quality binary segmentation (HQBS), including HRSOD \cite{hrsod}, UHRSD \cite{uhrsd}, DIS5K \cite{dis} and BIG \cite{cascadepsp}. 
We comprehensively compare these datasets in several important attributes for high-resolution segmentation tasks, including dataset size, spatial size, number of categories and mask complexity. Inspired by \cite{dis}, we also use the mean isoperimetric inequality quotient (mIPQ) \cite{ipq} metric to measure the overall complexity of masks in an image: $\mathrm{mIPQ}=\frac{1}{4 \pi n}\sum_{i=1}^{n}\frac{L_i^2}{A_i}$, where $L^i$ and $A_i$ denote the mask perimeter and the region area for the $i$th category, and $n$ denotes the total number of categories in this image.

The statistics are presented in \cref{tab:1}.
We can see that our dataset has a much higher average diagonal length, which can well represent image resolution, than existing semantic segmentation datasets. Higher resolution allows clearer and more detailed edges to be discerned. 
While the HQBS datasets also have high resolution, our dataset surpasses them in the number of images and the number of object categories. In addition, MaSS13K has a very small variance in image resolution, indicating a more consistent image quality.
In terms of the mIPQ metric, our dataset is 20 times more complex than the EntitySeg dataset, which is known for its high-quality annotations. Even compared to the most finely labeled HQBS dataset DIS5K, our MaSS13K still exceeds it by a significant margin.

%% file: sec/4_method.tex
\section{High-Resolution Segmentation Model}
\label{sec:method}

\input{figs/method_fig}

\subsection{Overview of MaSSFormer}

Our segmentation model, namely MaSSFormer, is built upon the architecture of Mask2Former \cite{mask2former}, as illustrated in \cref{fig:method}. The network consists of three main components: a pixel encoder, a pixel decoder, and a transformer decoder. Specifically, for an input image 
$I\in\mathbb{R}^{3 \times H \times W}$ with height $H$ and width $W$, the pixel encoder generates pixel embeddings $\{F_i\}_{i=1}^{4}$, which are then fed into the pixel decoder to produce multi-scale mask features $\{D_i\}_{i=1}^{4}$. 
Multi-scale mask features are used to update learnable queries through the Transformer decoder. The updated queries are then passed through  MLP layers to generate mask embeddings and class embeddings, which are combined with $D_1$ to produce the final semantic segmentation results.
In this work, we focus on the design of pixel decoder to accommodate high-resolution inputs and generate high-quality mask prediction with precise details.

In our pixel decoder, we use SE block with channel-attention \cite{senet} to squeeze the dimension of input pixel embeddings $\{F_i\}_{i=1}^{4}$ and obtain squeezed features $\{S_i\}_{i=1}^{4}$ of resolution $\{\frac{H}{k}\times\frac{W}{k},k=4,8,16,32\}$.
There are two key challenges for high-resolution and accurate semantic segmentation. First, for high-resolution inputs, the receptive field becomes relatively small, making the \textit{aggregation of high-level semantics difficult}. Second, \textit{precise object edges and fine details} are difficult to extract accurately. 
To address these challenges, we divide $\{S_i\}_{i=1}^{4}$ and image $I$ into two groups: $\{S_2,S_3,S_4\}$ and $\{S_1,I\}$, and aggregate these features with two parallel branches to obtain global semantics and local details, respectively.
In the first branch, we use a Cross Semantic Transmission (CST) module to aggregate global context while generating features with larger spatial resolution. We further introduce a Receptive Field Broaden (RFB) module to expand the receptive field while capturing multi-scale information. 
In the second branch, we introduce a Low-level Structure Extraction (LSE) module to directly capture high-resolution features with details.  Finally, an Edge Guided Fusion (EGF) module is employed to fuse the outputs of the two branches and efficiently produce high-resolution features for mask prediction. 


\subsection{Network Design Details}

\textbf{Global Semantic Branch}. As shown in \cref{fig:method}, we design a CST module to aggregate global semantic information. Inspired by \cite{pspnet,pem}, we apply global average pooling and a $1 \times 1$ convolution layer on $F_4$, and add it to $S_4$ to enhance the global context. Then, the context-enhanced features are fed into a deformable convolution layer with batch normalization and ReLU activation to obtain feature $D_4$, which contains global semantics but with a small spatial size. In contrast, $S_3$ shares similar semantics with $S_4$ while offering higher spatial resolution. We use Cross-Attention to transfer the detailed information from $S_3$ to $D_4$, followed by a Feed-Forward Network (FFN) $\mathrm{FFN(\cdot)}$. To reduce the computational cost, attention is performed within non-overlap windows. After passing through FFN, Self-Attention is computed within the window to further enhance semantic information at higher resolutions. Finally, another FFN is applied, followed by a deformable convolution to expand the receptive field. The whole process can be expressed as:
\begin{equation}
    \hat{D_4}=\mathrm{FFN}(\mathrm{CAttn}(\mathit{f_q}(\mathrm{US_2}(D_4)),\mathit{f_{k,v}}(S_3))),
\end{equation}
\begin{equation}
    D_3=\mathrm{DCN}(\mathrm{FFN}(\mathrm{SAttn}(\mathit{f_{q,k,v}}(\hat{D_4})))),
\end{equation}
where $\mathit{f_{\{qkv\}}}$ is linear projection and $\mathrm{US_n}(\cdot)$ means up-sampling with factor $n$.
By computing on low-resolution features, CST can gather global semantics at a lower computational cost and expand to larger spatial resolutions.

We then up-sample the CST output feature $D_3$ and add it to $S_2$ before feeding them to the RFB module for higher resolution semantic features $D_2$:
\begin{equation}
    \label{eq:RFB}
    D_2=\mathrm{RFB}(\mathrm{US_2}(D_3)+S_2).
\end{equation}
As shown in \cref{fig:method}, the RFB module includes 4 parallel deformable convolution modules with different kernel sizes  $[1,3,5,7]$ and a $1 \times 1$ point-wise convolution. 
What's more, RFB can capture multi-scale structure by setting convolution modules with different kernel size.

\noindent\textbf{Local Structure Branch}.
Traditional semantic segmentation methods \cite{deeplabv3+,mask2former,maskformer} typically do not consider high-resolution features for computational efficiency, which is unsuitable for high-resolution semantic segmentation. In this paper, we design the LSE module to extract low-level structure information with low computational cost. To generate high-quality masks, we upsample $S_1$ with resolution $\frac{H}{4} \times \frac{H}{4}$ to $\frac{H}{2} \times \frac{H}{2}$ and incorporate the down-sampled image to directly capture low-level features. However, the resolution of $\frac{H}{2} \times \frac{H}{2}$ results in a quadratic computational burden. As shown in \cref{fig:method}, we split the input into two sets of features with fewer channels, each concatenated with the image to learn distinctive features. We then use spatial attention \cite{cbam} to activate and extract low-level structural information. Finally, the features are concatenated and passed through a $1 \times 1$ convolution to output high-resolution features $S_{detail}$.

\noindent\textbf{Edge Guided Fusion}.
To reduce the cost in fusing low-level structure features at resolution $\frac{H}{2} \times \frac{W}{2}$, we design an EGF module to efficiently aggregate detailed information. First, we apply consecutive $1 \times 1$ convolutions to predict edges  $P^{edge}\in\mathbb{R}^{1\times H \times W}$ as follows:
\begin{equation}
    P^{edge} = \mathrm{Conv}(\mathrm{CBR}(S_{detail}+D_2)),
\end{equation}
where $\mathrm{Conv}$ means convolution and $\mathrm{CBR}$ means convolution with batch-norm and ReLU. During training, the edge detection task forces the network to learn the low-level structures of the image, allowing $S_{detail}$ to receive feedback and focus on edge regions. Then we use $S_{detail}$ to refine $D_2$, which can be expressed as:
\begin{equation}
\label{eq:refine}
    D_2^{refine} = \mathrm{Sigmoid}(S_{detail}) \times D_2.
\end{equation}
As shown in \cref{fig:method}, after reducing the channels with a $1 \times 1$ convolution, we apply a deformable convolution and restore the expanded channels with another $1 \times 1$ convolution. Finally, a residual connection is used to output  high-resolution features $D_1$. The intermediate features $\{D_2,D_3,D_4\}$ are then fed into the transformer decoder.

\noindent\textbf{Loss Function}.
\label{sec:loss}
We adopt the loss of Mask2Former \cite{mask2former}, which includes classification loss $L_{cls}$ and mask loss $L_{mask}$ (a combination of BCE loss and Dice loss). We assign an extra weight in the mask loss to highlight supervision on edge regions. The weights map for each target label $i$ can be generated using ground-truth label by $W^i=G^i-\mathit{f}_{avg}(G^i,k)$, where $\mathit{f}_{avg}(.,k)$ is an averaging filter with kernel size $k$ and $G^i$ is the $i$-th binary ground truth label. We assign the weights for the BCE loss as $L_{{BCE}}^{w} =\frac{1}{mn} \sum_i^n\sum^{m}_{j}l_{bce}(P_{j}^i,G_{j}^i)(1+\lambda W_{j}^i)$, where $i,j$ represent the $j$-th pixel on the $i$-th map and $\lambda$ is a hyperparameter that adjusts the weight ratio.
 Additionally, we incorporate an edge detection task in the BCE loss to supervise the edge prediction results, denoted by $L_{edge}=\frac{1}{n}\sum^n_i l_{bce}(P^{edge}_{i},G^{edge}_i)$, where $G^{edge}$ represents the edges between different semantic regions in the ground truth. The overall loss function is formulated as:
 \begin{equation}
     L_{total}=L_{BCE}^w+L_{Dice}+L_{cls}+L_{edge}.
     \label{eq:loss}
 \end{equation}

\subsection{Segmentation on New Classes}
Though our MaSS13K detaset consists of six commonly used categories and an `others' category, the `\texttt{others}' category actually contains various well segmented but unnamed objects. To fully exploit the precise segmentations of those objects in our dataset, we present a novel pipeline to enable MaSSFormer to segment higher-quality masks for new classes beyong the six predefined categories. 

We start with the image training set $\{I_i\}_{i=1}^n$ and the corresponding annotation set for 7 classes $\{G_i^j\}_{j=0}^6$, where $G_i^j$ represents the mask of the $i$-th image in the $j$-th class, and $\{G^0\}$ denotes the annotation set for `\texttt{others}'.
Using existing semantic segmentation models \cite{sam,grounded}, we can automatically annotate a new class in $\{I_i\}_{i=1}^n$ to generate a set of pseudo-labels, denoted by $\{\hat{G}^7_i\}$. By mixing the precise labels in the MaSS13K dataset and the pseudo-labels, we obtain a new set of labels to train the model.
We then design a label decoupling strategy, which employs the weighting scheme described in \cref{sec:loss} to treat the labels differently. 
Specifically, for the precise labels annotated in MaSS13K, we emphasize the loss on edges to update the network. For pseudo-labels, we apply inverse weights to ignore potentially erroneous edge annotations, allowing the network to focus on learning the main regions of the new classes. The training loss can be expressed as follows: 
\begin{equation}
\begin{split}
    L_{{BCE}}^{new} =\sum\nolimits_j^{1-6}(1+\lambda_1 W^j)L_{BCE}(P^j,G^j)+\\
    \sum\nolimits_j^{0,7}(1-\lambda_2 W^j)L_{BCE}(P^j,G^j),
\end{split}
\end{equation}
where $W^j$, $P^j$, $G^j$ represent weights map, predicted label and ground-truth label for the $j$-th class. In this way, we can train MaSSFormer to segment new classes one by one.




%% file: figs/method_fig.tex
\begin{figure*}[t]
    \centering
    \includegraphics[width=0.95\textwidth]{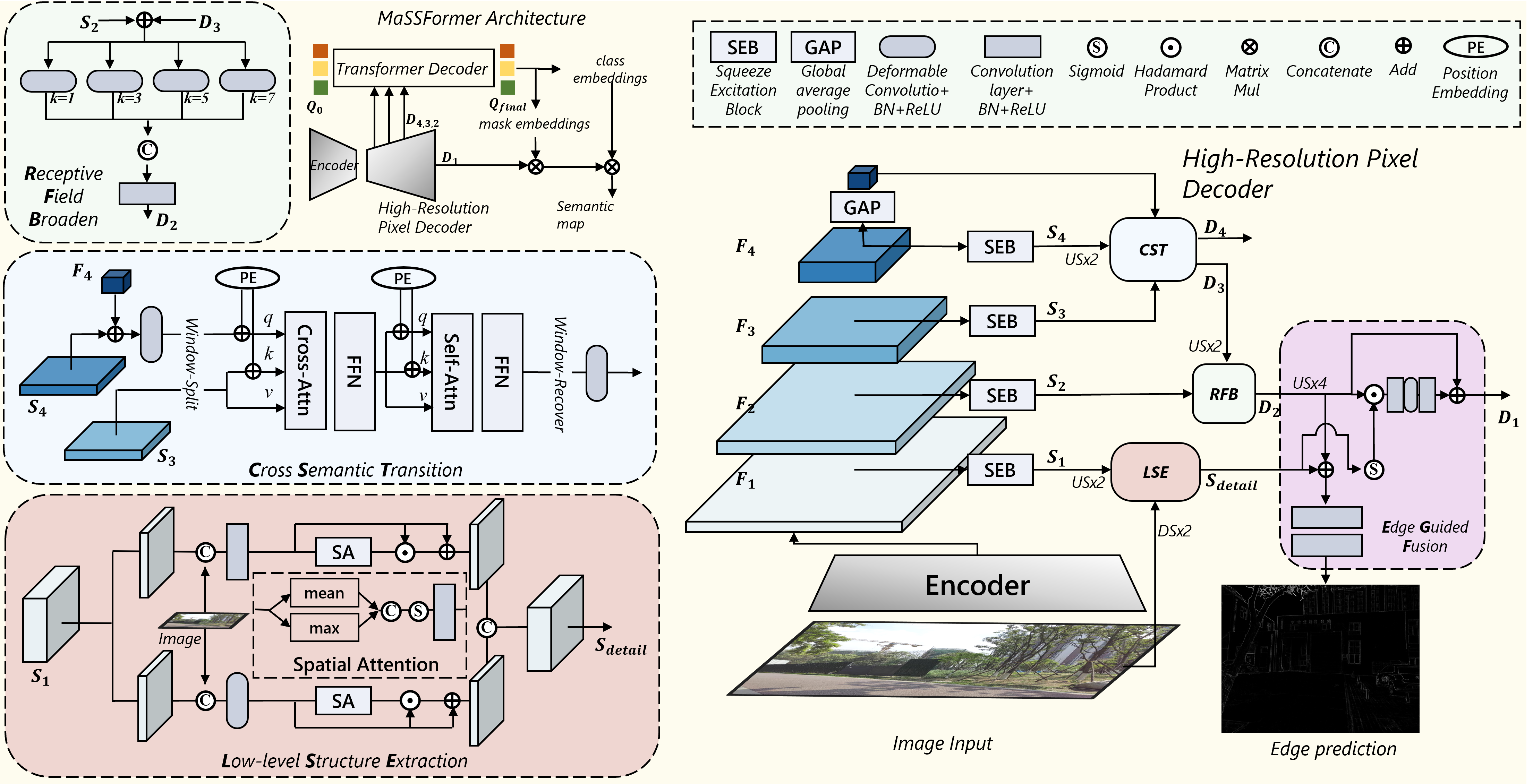}
    \caption{Architecture of MaSSFormer. The model architecture is shown on the top left corner, including the image encoder, pixel decoder and transformer decoder. The detailed structure of the high-resolution pixel decoder is shown on the right of this figure. }
    \label{fig:method}
    \vspace{-3mm}
\end{figure*}

%% file: sec/5_experiments.tex
\input{tables/main_results}
\input{figs/main_vis}

\section{MaSS13K Benchmark}
\noindent \textbf{Experiment Settings.} We present a comprehensive benchmarking on our MaSS13K dataset with existing semantic segmentation models and our MaSSFormer. We split the 13,348 images in MaSS13K into three subsets: \textbf{MaSS-train (11,348)}, \textbf{MaSS-val (500)}, and \textbf{MaSS-test (1,500)} for model training, validation, and testing, respectively.  
We select 14 representative and advanced semantic segmentation methods in the evaluation, which can be categorized into lightweight methods (including STDC2 \cite{stdc}, BeSeNetV2 \cite{bisenet}, PIDNet \cite{pid}, FeedFormer \cite{feed}, SegNext \cite{segnext}, SeaFormer \cite{sea} and CGRSeg \cite{cgrseg}), FCN-based methods (including DeepLabv3+ \cite{deeplabv3+}, UperNet \cite{upernet} and OCRNet \cite{ocrnet}), and Transformer-based methods (including MaskFormer \cite{maskformer}, Mask2Former \cite{mask2former}, PEM \cite{pem} and MPFormer \cite{mpformer}). We provide two variants of our MaSSFormer: the baseline version MaSSFormer with ResNet-50 backbone and the lightweight version MaSSFormer-Lite with ResNet-18 backbone. 
We implement MaSSFormer by mmsegmentation \cite{mmseg2020}. We use $1024\times1024$ crops during training and use the original resolution during testing for all methods. 
For the competing methods, we adopt their default settings in training on our MaSS13K. More training details of our MaSSFormer can be found in \textbf{supplementary materials}.

\noindent \textbf{Evaluation Metrics}.
Besides mIoU, which measures the overall segmentation accuracy, we use BIoU \cite{biou} and boundary F1-score \cite{bf1} to evaluate the boundary accuracy and quality for high-resolution semantic segmentation. More details can be found in the \textbf{supplementary materials}.

\subsection{Quantitative Evaluation}
\cref{tab:quantitative} shows the quantitative comparison between the proposed MaSSFormer and previous state-of-the-art semantic segmentation methods. One can see that MaSSFormer achieves the best results in terms of all the mIoU, BIoU and BF1 metrics while maintaining relatively low computational cost and parameter size, demonstrating its effectiveness and efficiency for high-resolution image segmentation. It is generally observed that correct segmentation of the main body is essential for accurate boundary segmentation, hence BIoU and BF1 tend to improve along with mIoU. However, some methods, such as FeedFormer and SegNext, show significant differences (+2.13\% on MaSS13K-val) in BIoU despite having similar (-0.54\%) mIoU scores. In addition, some segmentation errors in details edges and boundaries will have a small impact on the overall metrics like mIoU, but will significantly affect BIoU
 and BF1 scores, which can reflect a method's capability of segmenting object details. Our MaSSFormer achieves +0.69\% higher in mIoU and +1.57\% higher in BIoU than the second-best method, respectively, with the same ResNet50 backbone. Meanwhile, MaSSFormer with ResNet18 backbone also demonstrates competitive performance for high-resolution image segmentation, even higher than many networks with ResNet50 backbone, but with much fewer parameters and Flops. The results of MaSSFormer on each category can be found in the \textbf{supplementary materials}.

\subsection{Qualitative Evaluation}
\cref{fig:visual} presents qualitative comparisons between our MaSSFormer and three representative methods: Mask2Former, SegNext,  and DeepLabV3+, which have the best overall mIoU among the competing methods. We can see that the current methods struggle to distinguish fine details in high-resolution images, often resulting in blurry, discontinuous, or even missing segments. While this may have only a small impact on the mIoU and other metrics, it significantly degrades the visual quality and user experience of the segments. In contrast, our MaSSFormer effectively captures the details, producing higher-quality segmentation results.
Besides, MaSSFormer demonstrates high precision in various categories, including meticulous tower structures (top two rows) and thin branches and lines (bottom two rows).

\begin{figure}[t]
    \centering
    \includegraphics[width=0.9\linewidth]{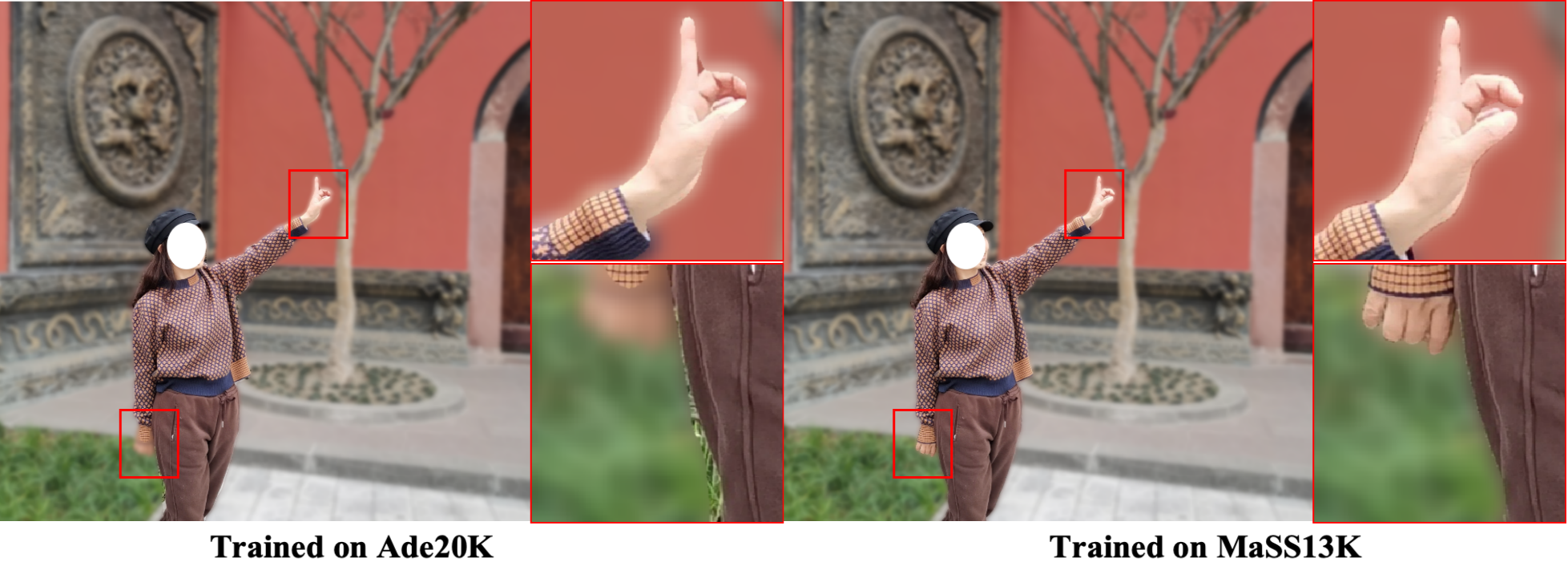}
    \vspace{-0.3cm}
    \caption{Comparison of bokeh effects with different masks. Please zoom-in for a better view.}
    \vspace{-0.2cm}
    \label{fig:bokeh}
\end{figure}

\noindent\textbf{Real-world Application.}
High-precision semantic image segmentation has many practical real-world applications. For example, mobile photography relies on accurate segmentation of portrait areas to achieve realistic bokeh effects. As shown in \cref{fig:bokeh}, the bokeh effect generated using the mask predicted by the model trained on MaSS13K is more natural and realistic than that by using the model trained on ADE20K, especially in the areas of the hands and legs. 
This demonstrates the importance of high-precision segmentation, which depends on high-quality and high-resolution datasets such as MaSS13K.

\subsection{Segmentation on New Class}
\begin{figure}[t]
    \centering
    \includegraphics[width=0.85\linewidth]{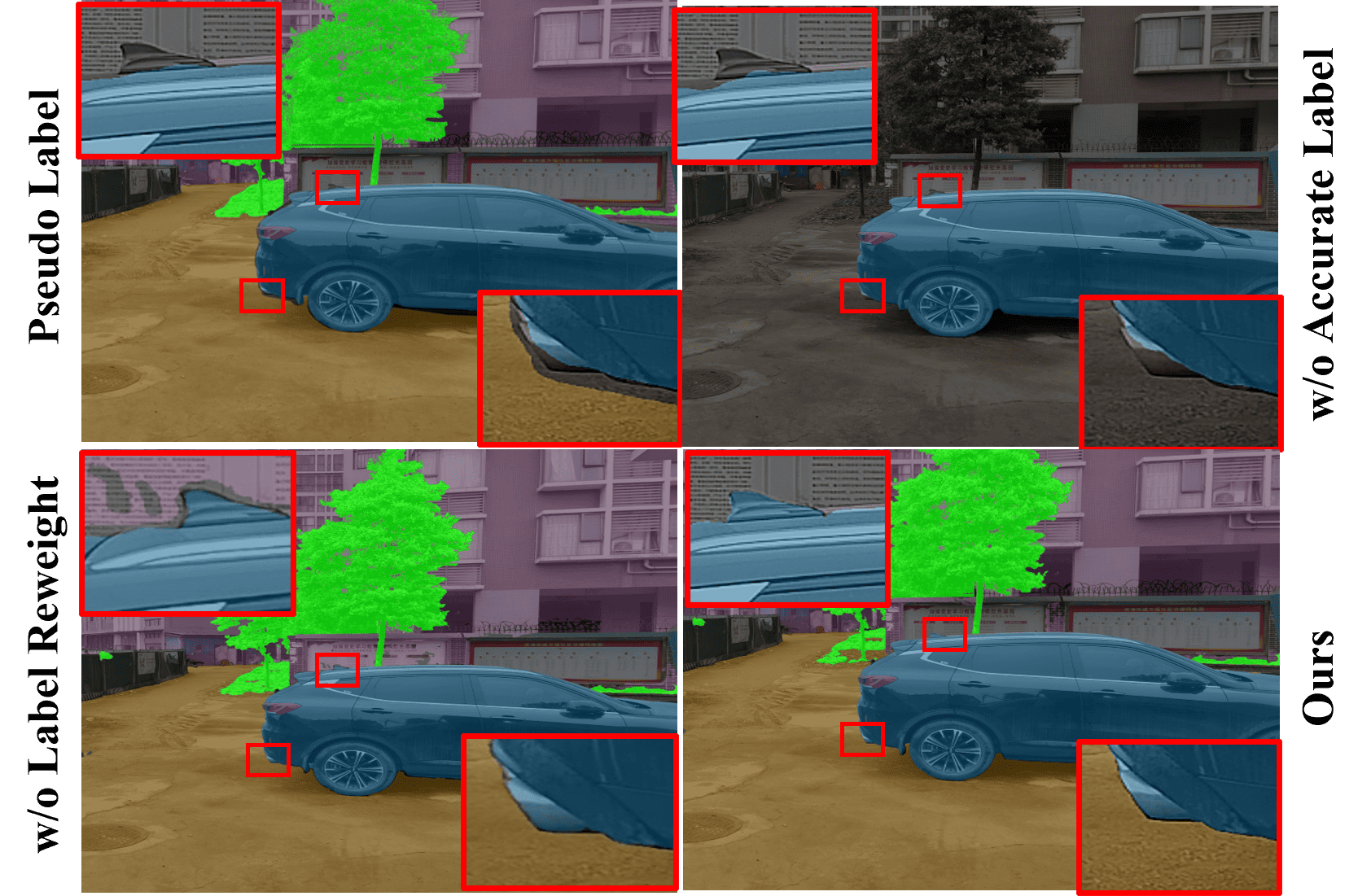}
    \vspace{-2mm}
    \caption{Visual results of MaSSFormer on novel class \texttt{Car}.}
    \vspace{-3mm}
    \label{fig:newclass}
\end{figure}

\cref{tab:newclass} and \cref{fig:newclass} show the segmentation results of MaSSFormer on a new class, `\texttt{Car}'. The pseudo-labels generated by existing semi-automatic tools achieve high mIoU but struggle with precise edge segmentation, especially when the target and surrounding areas are visually similar. Besides,  direct training with these pseudo-labels (upper-right of \cref{fig:newclass}) does not improve segmentation quality.
The bottom two rows of \cref{tab:newclass} show the results by mixed training with precise labels. We see that the model learns to enhance edge segmentation from precise annotations, improving the quality of car edges (see the bottom-left of \cref{fig:newclass}), thus increasing BIoU. However, while the car roof is accurately segmented, the surrounding regions are mistakenly classified as the `others' class because the incorrect pseudo-labels in the edge region can misguide the model training, leading to a 10.95\% decrease in mIoU (see the 3rd row of \cref{tab:newclass}).
With our re-weighting strategy, we reduce the weight of mislabeled edges, forcing the network to learn edge-aware capabilities from accurately labeled categories and new category features with higher reliability, thus achieving high-quality segmentation of new classes.
More experiments can be found in the \textbf{supplementary materials.}

\begin{table}[t]

\centering
 \renewcommand{\arraystretch}{0.8}
\caption{Quantitative evaluation on novel class \textbf{\texttt{Car}}.}
\vspace{-3mm}

\label{tab:newclass}

\resizebox{0.6\linewidth}{!}{%
\begin{tabular}{l|ll}
\toprule
 \multirow{2}{*}{Settings}  &\multicolumn{2}{c}{\textbf{\texttt{Car}}} \\ \cline{2-3}
 &\small {mIoU}  & \small{BIoU}     \\ \hline
 Pseudo Label &\small 94.18 &\small 20.44  \\
 \hline
 \small w/o Accurate Label&\small 92.43 &\small 22.52   \\
 \small w/o Label Re-weight &\small 83.23&\small 31.81\\
 \rowcolor{lightgray} \small w Label Re-weight &\small 95.21&\small 35.68    \\
 \bottomrule
\end{tabular}%
}
\vspace{-4mm}
\end{table}

\subsection{Ablation Study}
We conduct a series of ablation studies to validate the effectiveness of each component of MassFormer. 

\noindent\textbf{Global Semantic Aggregation.}
The top 3 rows in \cref{tab:2} demonstrate the effectiveness of our global semantic branch. By introducing the CST module, the mIoU metric is improved by 2.57\%, indicating that the acquisition of global semantics can enhance overall precision. Meanwhile, the increase in computational cost and parameters is slight.
By incorporating the RFB module, the receptive field increases, leading to a further improvement in mIoU. Compared to mIoU, there is a more significant increase in BIoU, indicating that our RFB module effectively fuses high-resolution features while maintaining semantic accuracy.

\noindent\textbf{Local Detail Extraction and Fusion.}
Comparing Exps\#3 and 4 in \cref{tab:2}, we see that the LSE module significantly increases the BIoU metric, implying that more low-level details can enhance the segmentation accuracy of edge boundaries. 
However, there is a slight decrease in mIoU, suggesting that direct fusion of low-level features may mislead the higher-level semantics.
Exp\#5 demonstrates that our designed edge-guided fusion strategy effectively improves low-level structure and high-level semantic feature fusion, resulting in an increase in both mIoU and BIoU.

\begin{table}[t]
 \renewcommand{\arraystretch}{0.85}
\centering
\caption{Ablation on MaSS-val dataset. }
\vspace{-4mm}

\label{tab:2}
\resizebox{0.9\linewidth}{!}{%
\begin{tabular}{l|p{0.4cm}p{0.4cm}p{0.4cm}p{0.6cm}|ll|ll}
\toprule
 Exp. &CST  &RFB &LSE &EGF &mIoU & BIoU &Para &FLOPs   \\ \hline
 \texttt{\#1}& & & & &85.54 &42.69 &34.72 &1298 \\
 \texttt{\#2} &\Checkmark& & &&88.11 &44.42 &35.96 &1348 \\
 \texttt{\#3} &\Checkmark&\Checkmark &&&88.29 &45.23 &37.32 &1692 \\
 \texttt{\#4} &\Checkmark&\Checkmark &\Checkmark&&88.02 &47.36 &37.40 &1928 \\
\rowcolor{lightgray}\texttt{\#5}&\Checkmark &\Checkmark &\Checkmark&\Checkmark &88.97 &48.97 &37.42 &2036 \\
 \bottomrule
\end{tabular}%
}
\vspace{-4mm}
\end{table}

%% file: tables/main_results.tex
\begin{table*}[t]
 \renewcommand{\arraystretch}{0.9}
\centering
\caption{Quantitative evaluation on MaSS13K validation and test sets. The best and the second-best results are  highlighted in \textbf{bold} and in \underline{underlined} respectively. FLOPs are all calculated for an input resolution of $4096\times4096$.}
\label{tab:quantitative}
\resizebox{0.9\linewidth}{!}{%
\begin{tabular}{lcc|ccc|ccc|cc}
\toprule
 \multirow{2}{*}{Method} &\multirow{2}{*}{Ori.}& \multirow{2}{*}{Backbone} & \multicolumn{3}{c|}{\textbf{MaSS-val(500)}}
 &\multicolumn{3}{c|}{\textbf{MaSS-test(1,500)}} & \multicolumn{2}{c}{Model Stat.}\\ 
 \cline{4-11} &&
 &mIoU$\uparrow$ & BIoU$\uparrow$ &BF1$\uparrow$ & mIoU$\uparrow$ & BIoU$\uparrow$&BF1$\uparrow$ & Param & FLOPs \\ \hline
 STDC2 \cite{stdc} & CVPR21&- &83.08 &27.99 &.3334 &83.76&27.72 &.3332&12.30M &564G \\
 BiSeNetv2 \cite{bisenet} &IJCV21   & - &71.55 &25.05&.3182& 72.92&24.48&.3171 &3.35M &591G\\
 SegNeXt \cite{segnext}&NIPS22&MSCAN-B &87.71 &39.93&.4615&\underline{88.11}&39.45&.4596 &27.57M &1536G \\
 PIDNet-L \cite{pid}  &CVPR23  &- &82.28 &31.30&.3475&81.77&30.70&.3479 &37.08M &1653G \\
 FeedFormer \cite{feed}&AAAI23&lvt &87.17 &42.06&.4838&86.56 &41.07&.4789&4.65M &300G \\
 SeaFormer-L \cite{sea}&ICLR23&- &86.78 &38.61&.4498&87.36 &38.28&.4489&13.95M &303G \\
  CGRSeg-L \cite{cgrseg} &ECCV24&EFv2-L &81.29 &34.48&.4090&81.45&33.95 &.4071&35.64M &1536G \\
 \midrule
 DeepLabv3+ \cite{deeplabv3+}&ECCV18& R50    & 86.66 & 40.08&.4718&85.14&38.65 &.4678& 41.22M &8008G \\
 UperNet      \cite{upernet}&ECCV18 &R50  &82.03 &35.85&.4181&81.98&35.61 &.4170&64.04M &11373G\\
 OCRNet      \cite{ocrnet}&ECCV20 &R50     &86.99 &32.68&.3808&83.02&31.33&.3731  &36.52M   &7352G  \\
 \midrule
 MaskFormer  \cite{maskformer}&NIPS21&R50 &83.27 &38.61&.4399&83.22&37.90 &.4393&41.31M &2396G \\
 Mask2Former \cite{maskformer}&CVPR22& R50    & \underline{88.28} & {47.40}&{.5458}&88.00&{46.13} &{.5330}& 44.01M &3123G  \\

 MPFormer \cite{mpformer}&CVPR23&R50 &87.76 &\underline{47.81} &\underline{.5513}&87.18&\underline{47.17}&\underline{.5486}&43.9M &4155G \\
 PEM \cite{pem}&CVPR24&R50 &83.41 &40.51 &.4675&83.38&39.99&.4644&35.5M &1859G \\
 \midrule
 \rowcolor{lightgray}MaSSFormer-Lite      &-&R18 &87.11 & 45.35&.5137&86.13&43.28 &.5086&15.07M &771G \\
 \rowcolor{lightgray}MaSSFormer      &-&R50 &\textbf{88.97}  &\textbf{48.97}&\textbf{.5639}&\textbf{88.21}&\textbf{48.39}  &\textbf{.5593}&37.42M  &2036G \\

 \bottomrule
\end{tabular}%
}
\end{table*}

%% file: figs/main_vis.tex
\begin{figure*}[t]
    \centering
    \includegraphics[width=0.9\textwidth]{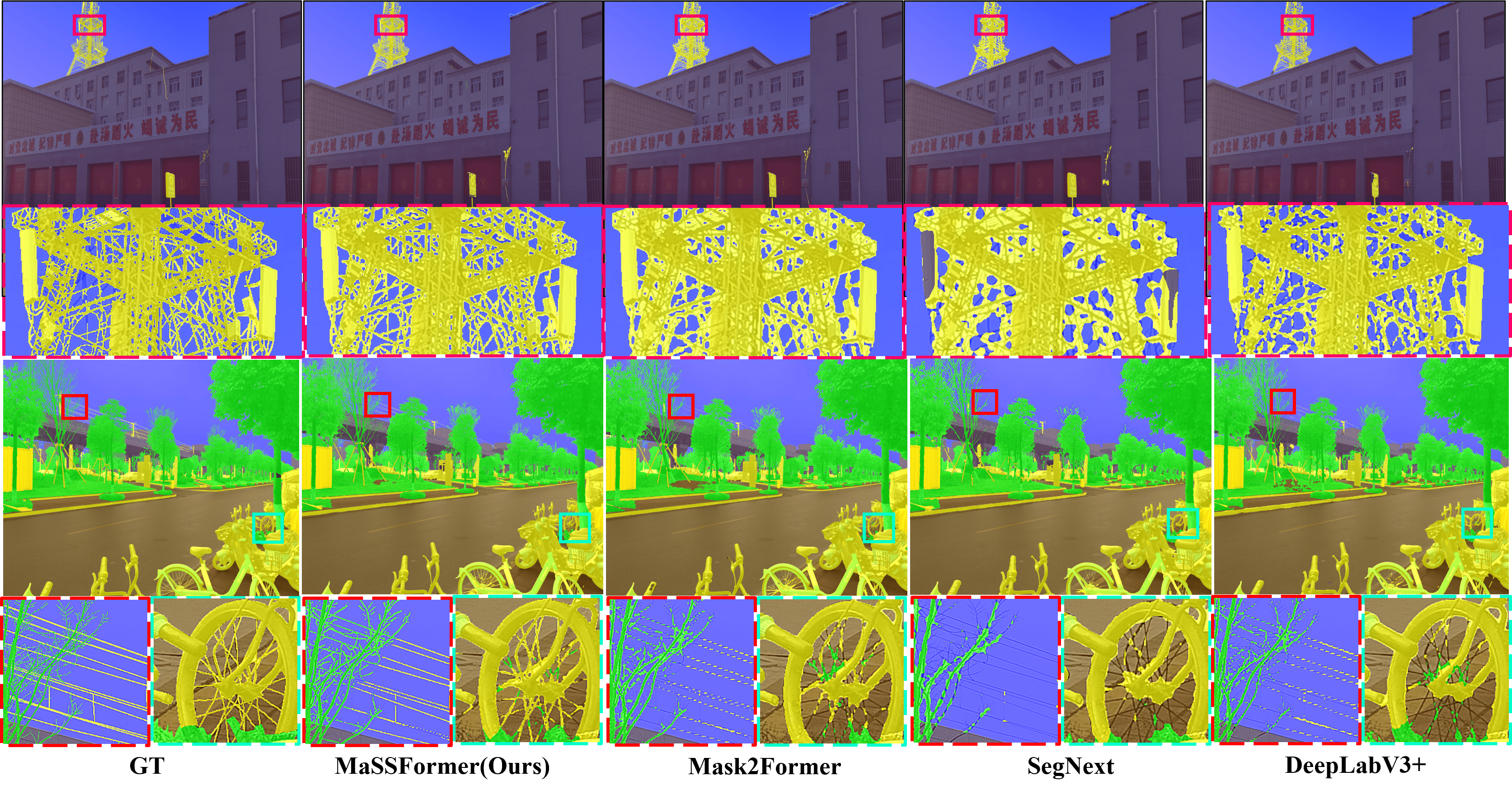}
\vspace{-3mm}   
 \caption{Qualitative comparisons of MaSSFormer with other three baseline methods. Please zoom-in for better view.}
    \label{fig:visual}
    \vspace{-3mm}
\end{figure*}

%% file: sec/6_conclusion.tex
\section{Conclusion}
\label{sec:conclusion}
In  this paper, we proposed MaSS13K and MaSSFormer for high-resolution semantic segmentation. The MaSS13K contained 13,348 real-word images at 4K resolution, with high-quality matting-level annotations in 7 categories. We then presented MaSSFormer, which efficiently aggregated global semantics and local structure details in high-resolution scenes, to address the challenges of high-resolution semantic segmentation.
We compared 14 representative methods with MaSSFormer on MaSS13K, establishing a comprehensive benchmark for high-resolution semantic segmentation. Furthermore, we proposed a scheme to transfer and generalize the fine segmentation capabilities of MaSSFormer to novel classes beyond the original categories, further revealing the potential value of MaSS13K dataset. We hope that MaSS13k can advance the research on high-resolution and high-quality semantic segmentation.

%% file: sec/X_suppl.tex
\clearpage

\begin{center}
    \Large \textbf{Supplementary Materials to \\ ``MaSS13K: A Matting-level Semantic Segmentation Benchmark"}
\end{center}
In this supplementary file, we provide the following materials:
\begin{itemize}
    \item More samples from our MaSS13K dataset (referring to Sec. 3 of the main paper).
    \item More details on training and evaluation metrics (referring to Sec. 5 of the main paper).
    \item More experimental results (referring to Sec. 5.1, 5.2, 5.3 of the main paper).
    \item Limitations.
\end{itemize}
\renewcommand\thesection{\Alph{section}}
\section{MaSS13K Dataset}
\subsection{More Samples from MaSS13K}

\begin{figure*}[b]
    \centering
    \includegraphics[width=0.7\textwidth]{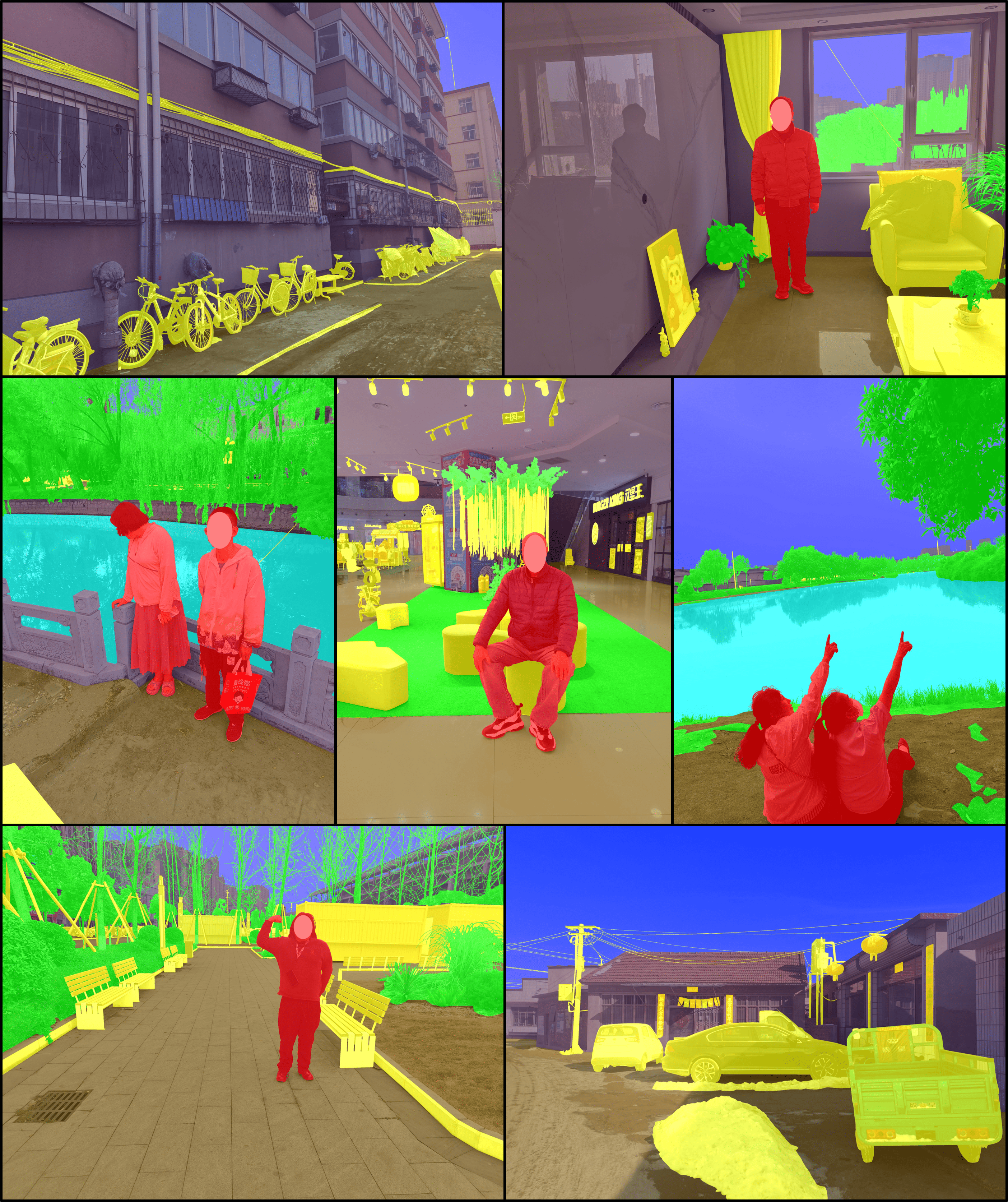}
    \caption{More samples from our MaSS13K dataset.}
    \label{fig:moresample}
\end{figure*}

We provide more annotated samples from our MaSS13K in \cref{fig:moresample}. MaSS13K covers a diverse range of indoor and outdoor scenes, such as urban areas, natural landscapes, street views, wilderness, parks, mall interiors, and other public spaces.

\subsection{Pixel Distribution of the Categories in MaSS13K}
\begin{figure}[t]
    \centering
    \includegraphics[width=0.8\linewidth]{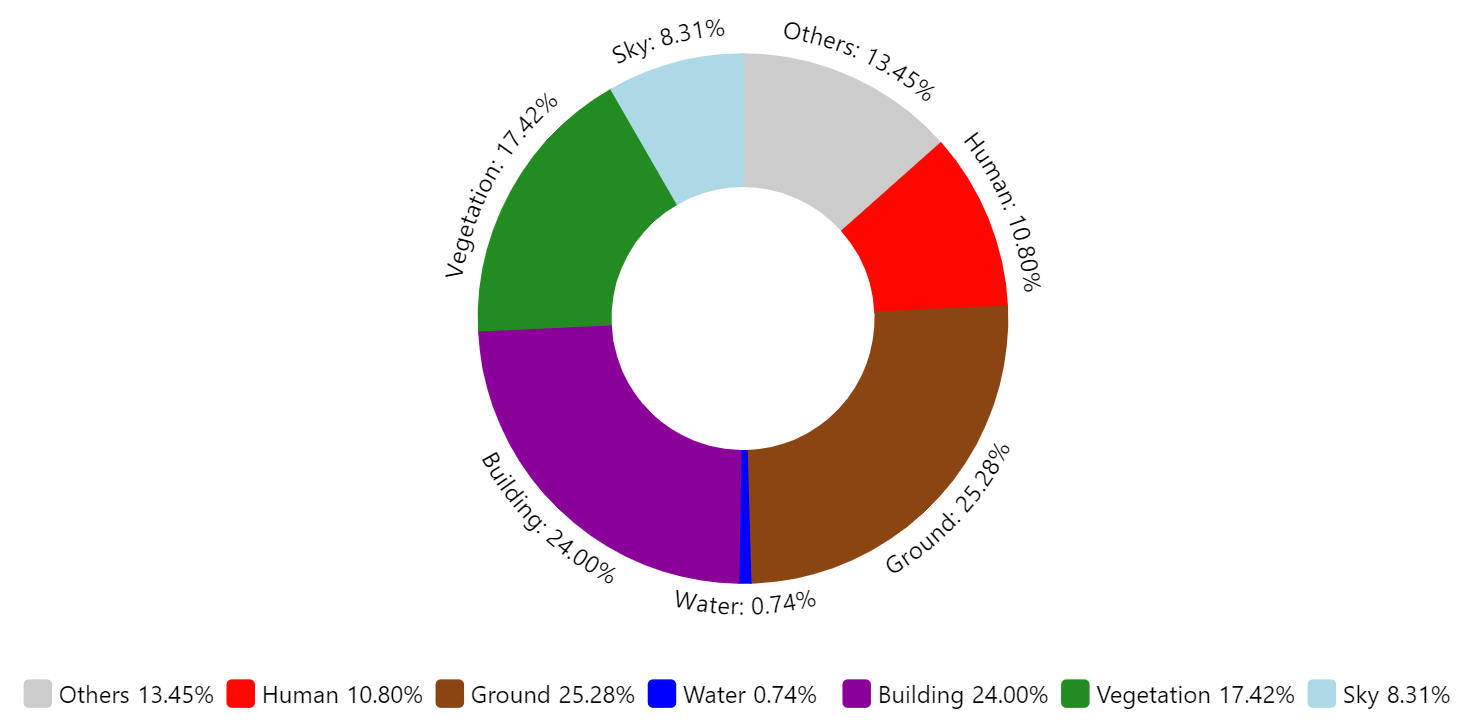}
    \caption{Pixel distribution of the seven categories in MaSS13K.}
    \label{fig:pie}
\end{figure}

\cref{fig:pie} illustrates the pixel distribution of the seven annotated categories in MaSS13K. Except for the ``\texttt{Water}" category, the pixel distribution of the other categories is relatively balanced. The dataset primarily consists of outdoor street scenes and urban landscapes, resulting in the highest pixel proportions for ``\texttt{Ground}" and ``\texttt{Building}". Additionally, 13.45\% of the pixels in the dataset are classified as ``\texttt{Others}", which includes various objects such as traffic signs, billboards, and vehicles. Despite being grouped into the ``\texttt{Others}" category, these objects are accurately labeled, ensuring the high quality of the dataset and its potential for further processing and development.

\section{More Details on Training and Evaluation Metrics}
\subsection{Training Details}
We implement MaSSFormer using mmsegmentation toolbox \cite{mmseg2020}. We use the AdamW \cite{adamw} optimizer to train our model with a batch size of 16.  The initial learning rate is set to 0.003 with a weight decay of 0.05, and the cosine decay schedule is applied during training.  
For data augmentation, we mainly follow the setup of Mask2Former \cite{mask2former}, including random resizing, random cropping, and random flipping. The models are trained for 80k iterations on the MaSS13K dataset. 

For the other evaluated methods on the MaSS13K dataset, we use their default settings on learning rate and data augmentation. The resolution during training and testing for all methods is kept consistent to ensure a fair comparison. 
The number of parameters and FLOPs for all methods are calculated using the mmsegmentation tools, except for PEM and MPFormer, which are calculated using Detectron2 tools.

\subsection{Details of Evaluation Metrics}
In high-resolution semantic segmentation, there are numerous fine-grained regions of object details that are critical for the quality of the segmentation masks. However, the standard mask IoU metric is too coarse to differentiate these fine-grained regions, making it less effective in evaluating high-resolution semantic segmentation performance. To better evaluate and compare the methods of high-resolution semantic segmentation, we also use boundary-focused metrics, including Boundary IoU (BIoU) \cite{biou} and Boundary F-1 Score (BF1) \cite{bf1}, in the main paper. 

\noindent\textbf{BIoU}. For the mask of the $i$-th category, the $\mathrm{BIoU}^i$ is defined as follows:
\begin{equation}
    \label{BIoU}
    \mathrm{BIoU}^i = \frac{(G_d^i \cap G^i)\cap (P_d^i \cap P^i)}{(G_d^i \cap G^i)\cup (P_d^i\cap P^i)},
\end{equation}
where $P$ and $G$ denote the predicted and ground-truth maps, respectively, and the subscript $d$ denotes the mask obtained by dilating the boundary by $d$ pixels. In our benchmark, we set $d$ to 0.1\% of the diagonal length, which is 5 pixels, to better measure the accuracy of details.

\noindent\textbf{BF1}. 
BF1 Score is a commonly used evaluation metric for edge detection and segmentation that combines precision and recall to assess the  edge quality of a segmentation map. The BF1 score is calculated as follows:
\begin{equation}
    \label{F1}
    \mathrm{BF1} = 2 \times \frac{\mathrm{Precision} \times \mathrm{Recall}}{\mathrm{Precision} + \mathrm{Recall}},
\end{equation}
where $\mathrm{Precision}=\frac{\mathrm{TP}}{\mathrm{TP}+\mathrm{FP}}$ and $\mathrm{Recall}=\frac{\mathrm{TP}}{\mathrm{TP}+\mathrm{FN}}$. In this case, if a  point of predicted boundary matches a ground truth boundary point within a distance error tolerance, it is considered a true positive. In our evaluation, we set the distance error tolerance  to 2. ``$\mathrm{TP+FP}$" can be represented by the total number of  edge points in the ground truth map, and ``$\mathrm{TP+FN}$" can be represented by the total number of edge points in prediction map.

\section{More Experimental Results}

\subsection{Ablation on Different Backbones}
In \cref{tab:bkbone}, we compare the performance of Mask2Former and MaSSFormer using stronger backbones. We  can observe that with stronger backbones, the IoU scores become higher due to the improved global semantic extraction capabilities. However, the BIoU and BF1 gaps between Mask2Former and our proposed MaSSFormer are enlarged, further validating that MaSSFormer can better process detail regions and produce higher-quality segmentation edges.
\begin{table}[h]
\renewcommand{\arraystretch}{1.5}
    \centering
    \resizebox{\linewidth}{!}{
    \begin{tabular}{l|l|lll|lll|ll}
    \toprule
         \multirow{2}{*}{Methods}&\multirow{2}{*}{Backbone} & \multicolumn{3}{c|}{\textbf{MaSS-val}}& \multicolumn{3}{c|}{\textbf{MaSS-test}} & \multicolumn{2}{c}{Stat.}\\
         \cline{3-10}
         &&mIoU & BIoU &BF1 & mIoU & BIoU &BF1  &Para. &FLOPS \\ \hline
         Mask2Former&Swin-T &89.17 &48.27 &0.5529 &89.47&47.97&0.5509 &47.40M&3148G \\ 
         \rowcolor{lightgray}MaSSFormer & Swin-T &89.49 \textbf{(+0.32)} &49.68 \textbf{(+1.41)} &0.5719 \textbf{(+0.0190)}&89.40 \textbf{(-0.07)}&49.50 \textbf{(+1.53)}&0.5685 \textbf{(+0.0176)} &40.92M&1956G \\ \hline
         
         Mask2Former & Swin-B &91.30 &51.20&0.5912 &91.10&50.83&0.5893 &107M&6126G \\ 
         \rowcolor{lightgray}MaSSFormer & Swin-B &91.35 \textbf{(+0.05)}&53.00 \textbf{(+1.80)}&0.6102 \textbf{(+0.0190)} &91.31 \textbf{(+0.21)}&52.74 \textbf{(+1.91)}&0.6076 \textbf{(+0.0183)} &100M&4890G \\
         \bottomrule
    \end{tabular}}
    \caption{Quantitative comparison with different backbones.}
    \label{tab:bkbone}
\end{table}
\subsection{Results of Each Category}
In \cref{tab:each_val} and \cref{tab:each_test}, we present detailed quantitative results for each category in \textbf{MaSS-val} and \textbf{MaSS-test}, respectively. From the two tables, it is evident that the `\texttt{Others}' class has the lowest IoU scores. This is because the `\texttt{Others}' class aggregates a variety of objects beyond the six categories, making it more challenging to identify and leading to lower overall IoU. In contrast, the `\texttt{Human}' and `\texttt{Sky}' categories are relatively distinct and well-defined, resulting in higher IoU scores. 
Additionally, we observe that MaSSFormer achieves the most significant improvement in BIoU for the `\texttt{Vegetation}' category compared to other methods. This aligns with our visual observations, as trees in the dataset contain a large amount of boundaries and details.

\begin{table*}[h]
 \renewcommand{\arraystretch}{1.5}
\centering
\caption{Quantitative evaluation on MaSS-val for each category.}
\label{tab:each_val}
\resizebox{1\linewidth}{!}{%
\begin{tabular}{l|ccc|ccc|ccc|ccc|ccc|ccc|ccc}
\toprule
 \multirow{2}{*}{Methods} & \multicolumn{3}{c|}{\texttt{Others}}
 &\multicolumn{3}{c|}{\texttt{Human}} & \multicolumn{3}{c|}{\texttt{Building}}& \multicolumn{3}{c|}{\texttt{Vegetation}}& \multicolumn{3}{c|}{\texttt{Ground}}& \multicolumn{3}{c|}{\texttt{Sky}}& \multicolumn{3}{c}{\texttt{Water}}\\ 
 \cline{2-22} 
 &mIoU & BIoU &BF1&mIoU & BIoU &BF1&mIoU & BIoU &BF1&mIoU & BIoU &BF1&mIoU & BIoU &BF1&mIoU & BIoU &BF1&mIoU & BIoU &BF1  \\ \hline
 STDC2 &66.52 &18.39 &.2041 &95.77 &45.06 &.5366&77.96 &19.78&.2467 &87.88 &22.84&.2701 &90.62 &32.54&.4311 &89.96 &34.21&.3647 &73.51 &23.66 &.2582\\
BiSeNetV2 &53.26 &13.31 &1589&92.38 &38.90 &.5087&69.93 &15.27&.2013 &83.62 &27.45&.3013 &85.16 &26.84&.3940 &82.90 &41.45 &.4637&35.83 &11.40 &.1417\\
SegNext &70.17 &25.37&.3050 &97.58 &59.61&.7061 &82.22 &27.68&3440 &89.30 &36.58&.4132 &93.74 &40.94&.5392 &92.40 &48.89&.5783 &88.60 &40.50&.4462\\
PIDNet-L &64.71 &19.83&.2258 &96.16 &44.74 &.4597&77.74 &21.82&.2620 &86.64 &29.74&.3278 &89.84 &32.24&.4324 &89.64 &41.85&.4515 &71.15 &28.90&.3144\\
FeedFormer &65.90 &23.34 &.2870&97.02 &57.68 &.6986 &79.46 &26.97 &.3465&89.67 &44.59 &.5006&92.66 &38.93 &.5187&94.10 &62.19 &.7321&91.42 &40.73&.4464\\
SeaFormer &69.75 &24.53&.2918 &97.10 &57.88&.7029 &82.63 &26.93&.3349 &88.41 &34.61&.3955 &92.16 &39.69&.5211 &93.59 &48.56&.5565 &83.83 &38.13&.4279\\
CGRSeg &59.32 &21.83 &.2618&90.42 &47.90 &.6129&74.05 &23.42 &.2917&88.15 &34.30 &.3843&87.47 &34.53 &.4699&88.88 &46.60 &.5486&80.80 &32.78&.4031\\
\hline
DeepLabV3+ &67.61 &22.76 &.2947&96.30 &56.36 &.7074&78.83 &25.30&.3359 &89.35 &41.46&.4831 &90.25 &34.66 &.4935&94.09 &60.35 &.6822&90.28 &40.38 &.4518\\
UperNet &62.60 &20.15 &.2554 &92.60 &45.40 &.5772 &77.32 &23.63 &.2941 &88.20 &39.85 &.4479 &88.79 &33.03 &.4553 &91.44 &58.36 &.6253 &73.46 &31.39 &3456 \\
OCRNet &61.11 &18.34&.2304 &93.91 &46.63&.6051 &75.17 &20.19&.2617 &88.22 &31.15&.3577 &89.35 &31.17 &.4398&92.45 &47.05&.5077 &87.78 &34.61&.3548\\
\hline
MaskFormer &60.16 &19.82 &.2680&94.55 &51.09&.6463 &72.94 &22.80 &.3043&87.83 &43.98 &.4844&85.69 &29.78 &.4303&93.96 &62.15 &.7159&87.72 &40.65 &.4194\\
Mask2Former &72.45 &31.23 &.3907&98.13 &66.08 &.7755&82.95 &32.86 &.4165&90.25 &45.99 &.5170&93.23 &44.72 &.5824&94.52 &63.59 &.7417&86.42 &47.35 &.5241\\
MPFormer &73.34 &31.84 &.3913&98.49 &67.49&.7885 &84.29 &33.93&.4258 &89.93 &46.90 &.5305&93.04 &44.68 &.5772&93.59 &64.29 &.7475&81.68 &45.54&.5139\\
PEM &68.10 &25.46&.3198 &97.50 &60.32&.7288 &70.49 &23.61&.3232 &89.75 &44.25&.4929 &77.71 &26.98&.4225 &93.31 &61.66&.7171 &87.02 &41.30&.4480\\
\hline

\rowcolor{lightgray} MaSSFormer-Lite &67.62 &26.02 &.3288&96.98 &61.01 &.7352&79.65 &30.01 &.3854&88.90 &49.31 &.5374&90.59 &38.90 &.5327&95.35 &68.26 &.7587&90.65 &43.97&.4784\\
\rowcolor{lightgray} MaSSFormer &71.27 &30.84&.3931 &97.44 &64.62&.7783 &82.74 &33.74&.4385 &89.76 &50.36&.5799 &93.40 &44.13 &.5605&95.90 &69.91 &.8108&92.32 &49.18&.5594\\

 \bottomrule
\end{tabular}%
}
\end{table*}

\begin{table*}[h]
 \renewcommand{\arraystretch}{1.5}
\centering
\caption{Quantitative evaluation on MaSS-test for each category.}
\label{tab:each_test}
\resizebox{1\linewidth}{!}{%
\begin{tabular}{l|ccc|ccc|ccc|ccc|ccc|ccc|ccc}
\toprule
 \multirow{2}{*}{Methods} & \multicolumn{3}{c|}{\texttt{Others}}
 &\multicolumn{3}{c|}{\texttt{Human}} & \multicolumn{3}{c|}{\texttt{Building}}& \multicolumn{3}{c|}{\texttt{Vegetation}}& \multicolumn{3}{c|}{\texttt{Ground}}& \multicolumn{3}{c|}{\texttt{Sky}}& \multicolumn{3}{c}{\texttt{Water}}\\ 
 \cline{2-22} 
 &mIoU & BIoU &mIoU &BF1& BIoU&mIoU &BF1& BIoU&mIoU &BF1& BIoU&mIoU &BF1& BIoU&mIoU &BF1& BIoU&mIoU &BF1& BIoU&BF1  \\ \hline
 STDC2 &66.75 &17.88&.2008 &94.98 &43.71&.5246 &78.97 &19.69&.2492 &87.72 &22.36&.2680 &90.35 &31.82&.4302 &91.20 &34.27 &.3712&76.37 &24.29&.2622\\
BiSeNetV2 &52.63 &12.84&.1600 &90.53 &37.35&.4967 &70.37 &15.24&.2010 &83.31 &26.91&.2979 &84.90 &26.45 &.3934&86.50 &41.27&.4625 &42.23 &11.27&.1500 \\
SegNext &69.80 &24.63&.3031 &97.41 &58.45&.6935 &82.40 &27.45&.3472 &89.41 &36.24&.4076 &93.49 &39.78&.5316 &93.50 &49.49&.5775 &90.76 &40.12&.4490\\
PIDNet-L &64.63 &19.65&.2269 &95.47 &43.50&.4505 &78.24 &22.13&.2668 &87.07 &29.38&.3210 &89.69 &31.77&.4304 &89.95 &42.08&.4586 &67.31 &26.38&.3191\\
FeedFormer &65.67 &22.47 &.2823&96.29 &55.91 &.6824&79.82 &26.70&.3464 &89.55 &44.06 &.4898&92.34 &37.80 &.5138&94.29 &62.28 &.7281&88.02 &38.31&.4457\\
SeaFormer &69.30 &24.23&.2927 &96.23 &56.49&.6923 &82.68 &27.04&.3373 &88.66 &34.30&.3883 &92.13 &38.77&.5172 &93.61 &48.96&.5556 &88.96 &38.23&.4354\\
CGRSeg &59.17 &21.09 &.2596&90.65 &46.80 &.6038&73.96 &23.37 &.2952&87.97 &33.82 &.3778&87.68 &33.78 &.4686&90.64 &46.92&.5426 &80.09 &31.92&.3942\\
\hline
DeepLabV3+ &67.32 &22.46 &.2930&95.65 &54.11 &.6906&79.25 &25.66&.3363 &89.35 &41.53&.4763 &89.76 &33.95&.4859 &93.37 &60.16 &.6816&81.25 &32.65&.4224\\
UperNet &61.97 &19.78 &.2515&90.70 &43.30&.5601 &77.74 &24.19&.3000 &88.20 &40.29&.4429 &88.20 &32.56&.4500 &92.19 &59.08 &.6318&74.87 &30.06&.3617\\
OCRNet &60.60 &17.57 &.2222&92.41 &43.44&.5875 &75.76 &20.08 &.2580&87.74 &30.36&.3437 &88.48 &29.83&.4306 &92.65 &46.95&.5070 &83.56 &31.10&.3501\\
\hline
MaskFormer &61.28 &19.65 &.2679&94.48 &50.26 &.6430&74.64 &22.99&.3059 &88.24 &44.14&.4818 &85.61 &29.04 &.4241&94.25 &62.80 &.7160&84.06 &36.45&.4182\\
Mask2Former &70.67 &29.60 &.3708&97.69 &65.02 &.7656&80.83 &31.94 &.4033&90.25 &44.37 &.4985&91.66 &42.60 &.5608&94.88 &64.13&.7241 &90.00 &45.27&.5233\\
MPFormer &72.38 &31.23&.3896 &98.06 &66.13&.7784 &83.91 &33.67&.4271 &90.17 &46.70&.5228 &92.58 &43.68&.5718 &93.57 &64.81&.7451 &79.63 &44.02&.5124\\
PEM &67.70 &25.04&.3210 &97.46 &59.97&.7205 &72.14 &24.02&.3287 &89.12 &44.01&.4854 &78.75 &26.12&.4146 &91.86 &61.25&.7094 &86.69 &39.51&.4593\\
\hline

\rowcolor{lightgray} MaSSFormer-Lite  &66.80 &25.11&.3260 &96.39 &59.19&.7216 &79.60 &29.55&.3844 &88.77 &46.75&.5280 &90.87 &37.69&.5170 &94.77 &64.54&.7548 &85.71 &40.08&.4707\\
\rowcolor{lightgray} MaSSFormer &70.07 &29.70&.3860 &97.23 &64.02&.7665 &81.65 &33.40&.4387 &90.03 &51.91&.5772 &91.51 &41.56&.5557 &95.60 &70.30&.8043 &90.18 &47.82&.5611\\

 \bottomrule
\end{tabular}%
}
\end{table*}

\subsection{More Details and Results on Segmenting New Classes}

\textbf{Details of the Pipeline}. We utilize the Grounded-SAM \cite{grounded} as a semi-automatic labeling tool to obtain pseudo labels without human effort. Grounded-SAM employs the bounding boxes from Grounding-DINO \cite{gdino} as the input to SAM \cite{sam} to generate masks for the corresponding objects. Grounding-DINO, as an open-vocabulary detection model, can take any text prompt and produce bounding boxes for objects of that category in the image. Therefore, for a new category `\texttt{X}', we input `\texttt{X}' as a text-prompt into Grounded-SAM, process all images in the dataset to obtain instance segmentation masks for that category, and then merge and convert these masks into semantic segmentation labels for the new category `\texttt{X}'.
We combine the semantic labels of the new category with the existing seven categories' labels for joint model training. To provide a quantitative evaluation of the mask quality for the new category, we manually annotate a set of images of that category from the test set to calculate the mIoU, BIoU, and BF1 scores. What's more, in the training on new categories, we uniformly use a crop size of 2048x2048.

\begin{figure}
    \centering
    \includegraphics[width=0.8\linewidth]{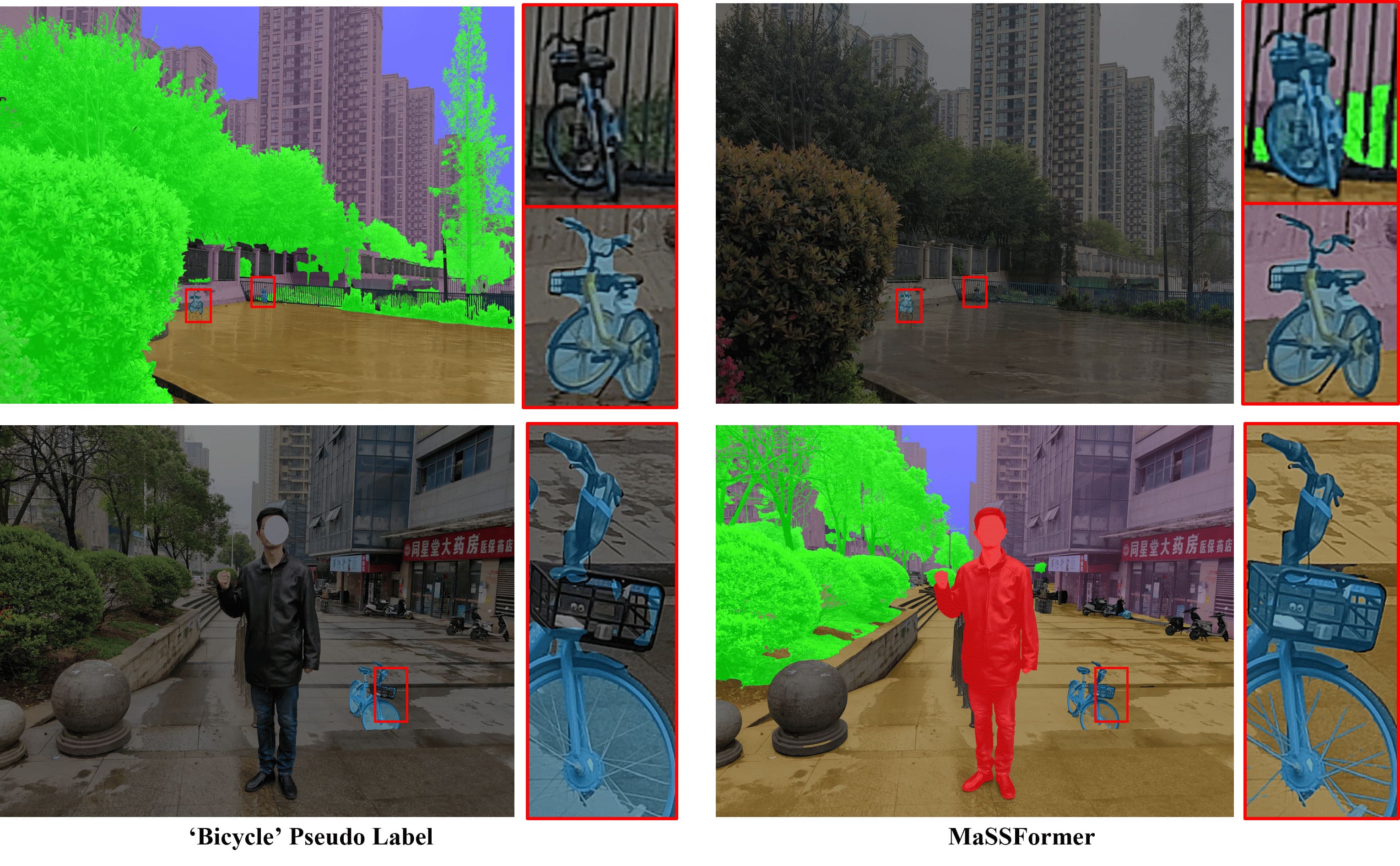}
    \caption{Visual results on the segmentation of `\texttt{Bicycle}' class. Left: Pseudo-labels generated by Grounded-SAM. Right: Predictions by MaSSFormer.}
    \label{fig:sup_bicycle}
\end{figure}

\noindent\textbf{More Results}. 
In the main paper, we have validated the effectiveness of our method in segmenting a new class `\texttt{Car}' on MaSS13K. In this supplementary file, we conduct experiments on another new class `\texttt{Bicycle}'. The results are shown in \cref{tab:sup_newclass}. For the convenience of readers, we also put the results of class `\texttt{Car}' in the table. We can see that for both of the two new classes, the IoU, BIoU, and BF1 metrics show significant improvements over the baseline. 
Some visual examples of the segmentation results of `\texttt{Bicycle}' are presented in \cref{fig:sup_bicycle}. We can see that due to the relatively small size of bicycle targets, the incorrect segmentation of the wheels can severely impact the IoU scores, as shown in the 2nd row of \cref{fig:sup_bicycle}. In addition, the failure of Grounded-SAM to detect small targets can further reduce the IoU, as illustrated in the 1st row of \cref{fig:sup_bicycle}. Our method, designed for high-resolution images, can effectively capture fine structural details and boundaries, resulting in higher IoU and BF1 scores. Furthermore, under the joint supervision of other precise categories, our method can accurately distinguish foreground from background at the wheel areas, resulting in precise segmentation of the target objects.



\begin{table}[t]

\centering
\caption{Quantitative evaluation on novel classes \textbf{\texttt{Bicycle}} and \textbf{\texttt{Car}}.}

\label{tab:sup_newclass}

\resizebox{0.8\linewidth}{!}{%
\begin{tabular}{l|lll|lll}
\toprule
 \multirow{2}{*}{Settings}  &\multicolumn{3}{c|}{\textbf{\texttt{Bicycle}}}&\multicolumn{3}{c}{\textbf{\texttt{Car}}} \\ \cline{2-7}
 & {mIoU}  & {BIoU} &BF1 & {mIoU}  & {BIoU} &BF1     \\ \hline 
 Pseudo label generated by Grounded-SAM &49.40 &23.82 &0.2800 &94.18 &20.44&0.2522  \\
 \hline
 \rowcolor{lightgray} Prediction generated by our joint-trained model &74.57 &40.17 &0.4609 &95.21 &35.68&0.3643    \\
 \bottomrule
\end{tabular}%
}

\end{table}

\subsection{More Qualitative Comparisons}
\begin{figure*}[h]
    \centering
    \includegraphics[width=\textwidth]{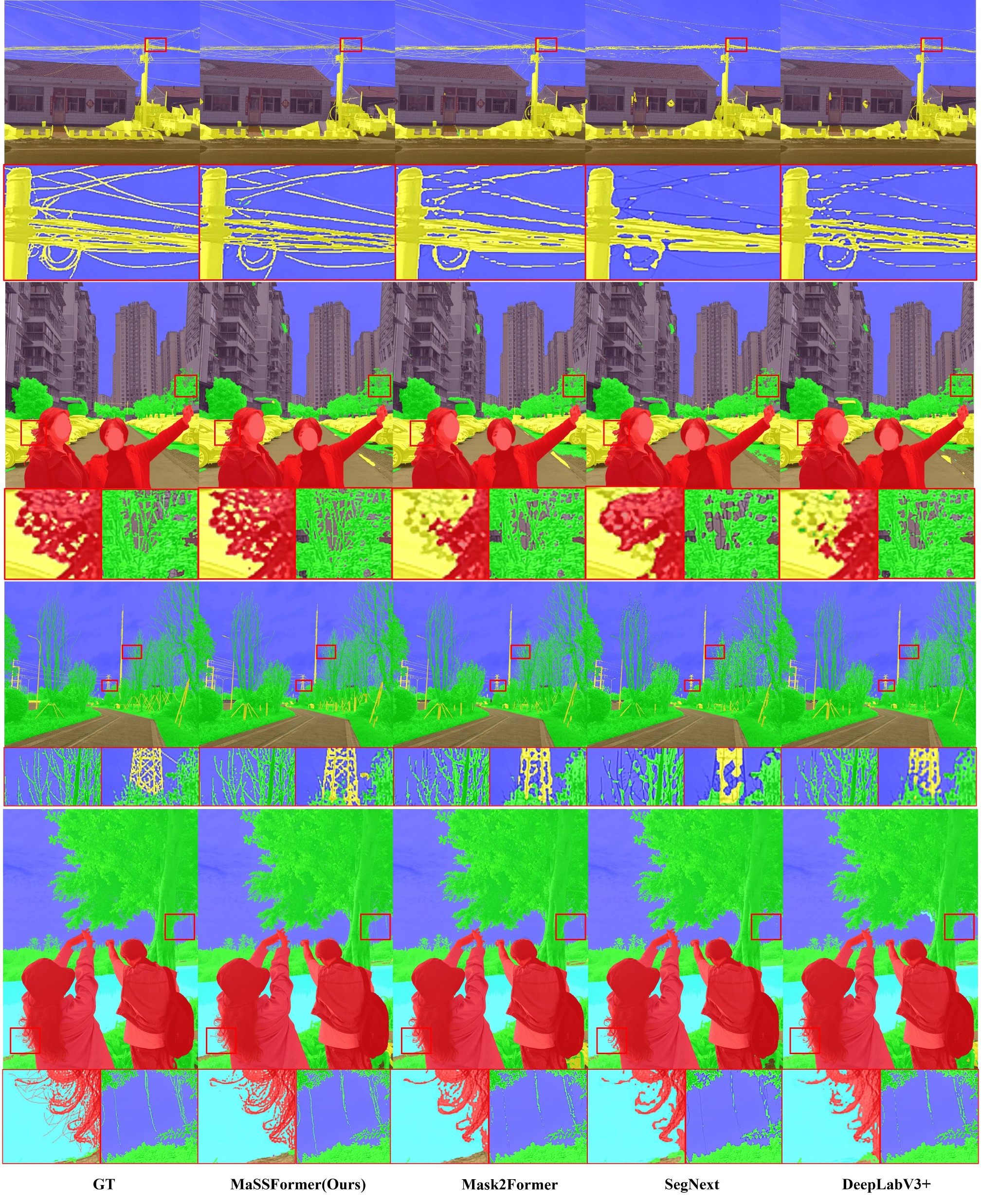}
    \caption{More qualitative comparisons between MaSSFormer and other methods. Please zoom-in for a better view.}
    \label{fig:sup_vis_comp}
\end{figure*}

We present more qualitative comparisons between our MaSSFormer and other representative methods in \cref{fig:sup_vis_comp}. It can be seen that MaSSFormer demonstrates superior performance in segmenting fine-grained regions, such as the thin  lines in the 1st image. It maintains accurate segmentation even for small objects in the distance (the 3rd image). For fine structures such as hair, the competing methods often fail to achieve fine-level segmentation and tend to predict the surrounding areas as hair (the 2nd and 4th images. In contrast, MaSSFormer effectively distinguishes hair and other detailed elements from the background, ensuring high-quality segmentation.

\section{Limitations} First, while MaSS13K provides 13K finely annotated images, it can be further expanded in the number of samples, scenes, and categories. Second, although MaSSFormer has achieved a relatively good balance between accuracy and efficiency, its computational cost and memory usage are still high, especially for mobile devices. New lightweight networks are expected for efficient yet accurate high-resolution semantic segmentation. 